\pdfoutput=1
\documentclass[11pt]{article}

\usepackage[]{acl}

\usepackage{times}
\usepackage{latexsym}

\usepackage[T1]{fontenc}
\usepackage[utf8]{inputenc}

\usepackage{microtype}

\usepackage{inconsolata}

\usepackage{utfsym}

\usepackage{multirow}
\usepackage{xcolor, colortbl}
\usepackage{graphicx,subcaption}
\usepackage{amsmath}

\title{LlamaTurk: Adapting Open-Source Generative Large Language Models for Low-Resource Language}

\author{Cagri Toraman \\
         Department of Computer Engineering\\
         Middle East Technical University, Ankara, Turkey\\ 
         \texttt{ctoraman@ceng.metu.edu.tr} 
         }

\begin{document}
\maketitle
\begin{abstract}
Despite advancements in English-dominant generative large language models, further development is needed for low-resource languages to enhance global accessibility. The primary methods for representing these languages are monolingual and multilingual pretraining. Monolingual pretraining is expensive due to hardware requirements, and multilingual models often have uneven performance across languages. This study explores an alternative solution by adapting large language models, primarily trained on English, to low-resource languages. We assess various strategies, including continual training, instruction fine-tuning, task-specific fine-tuning, and vocabulary extension. The results show that continual training improves language comprehension, as reflected in perplexity scores, and task-specific tuning generally enhances performance of downstream tasks. However, extending the vocabulary shows no substantial benefits. Additionally, while larger models improve task performance with few-shot tuning, multilingual models perform worse than their monolingual counterparts when adapted.
\end{abstract}

\section{Introduction}

\begin{figure}[t]
\centering
\includegraphics[width=0.35\textwidth]{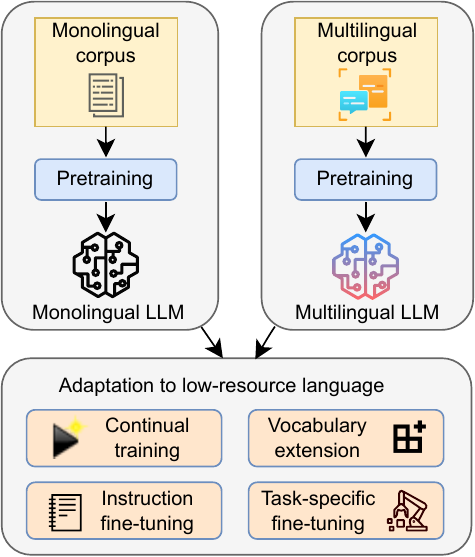}
\caption{Adapting generative large language models for low-resource languages.}
\label{fig:illustration}
\end{figure}

The performance of proprietary generative large language models (LLMs) is better than open-source ones in most cases as this article is written \cite{10.1145/3520312.3534862, sun2024trustllm}, though there are efforts to develop open-source generative LLMs in terms of high performance and human ethics alignment \cite{touvron2023llama, jiang2023mistral, almazrouei2023falcon}.

The progress is more significant in the English language compared to other languages as the aforementioned open-source models are mostly trained by English corpora \cite{wang2023languages, zhang-etal-2023-dont}. To make natural language processing technology more inclusive and accessible globally, research and development should be dedicated to the techniques that improve the performance of large language models in low-resource languages.

Monolingual \cite{yang2023mono,nagoudi2023jasmine,uludoğan2024turna,correa2024teenytinyllama,kesgin2024introducing} and multilingual pretraining \cite{shliazhko2023mgpt,bloom,lin2024mala500} of generative LLMs are two main solutions for representing low-resource languages. However, monolingual pretraining is too costly due to hardware requirements for generative LLMs \cite{zhao2023survey}. On the other hand, multilingual LLMs have uneven performance across different languages mostly due to imbalanced training corpus \cite{zhang-etal-2023-dont, qin2024multilingual}. Our proposed solution is to adapt open-source generative LLMs for low-resource languages, illustrated in Figure \ref{fig:illustration}.

In this regard, this study examines how to adapt open-source LLMs for low-resource languages in a systematic way. We focus on the benefits of using different methodologies, both individually and together, including continual training, supervised fine-tuning, and vocabulary extension, to adapt generative LLMs for low-resource languages. 

For the sake of efficiency, we use Llama \cite{touvron2023llama} in the experiments. We select the Turkish language as a low-resource language. %\footnote{We refer to the Turkish language as low-resource language in NLP, since existing open-source generative LLMs do not directly support its usage.}. 
We therefore refer to the model family used in this study as \texttt{LlamaTurk}. The model size and language selection are affordable when the number of experiments is considered in this study\footnote{Two NVIDIA RTX 2080Tis and four A4000s are employed in the experiments.}. Also, Llama is trained mostly with English data, which can provide better investigation for adapting non-English languages. The Turkish language can be categorized under low-resource languages when training corpus of open-source generative LLMs are considered \cite{touvron2023llama}, yet the recipes given in this study can also be used for other low-resource languages since the methods are independent of language itself. 

We further examine adaptation in terms of two more aspects: Model size and multilinguality. Model size is important for scalability and performance \cite{zhao2023survey,yang2023harnessing}. We provide an analysis of the adaptation of Llama-7b and 13b in this respect. Moreover, multilingual LLMs, such as BLOOM \cite{bloom}, Yi \cite{ai2024yi}, Aya \cite{aya2024}, and MaLA \cite{lin2024mala}, can provide an opportunity to adapt low-resource languages easier than English-dominant ones due to multilingual corpus and vocabulary. Since BLOOM and Yi do not involve Turkish in training and Aya is larger than MaLA in terms of model parameters, we use MaLA for an analysis of multilingual LLMs.

The main contributions of this study can be summarized as follows. We (i) analyze the adaptation of generative LLMs for low-resource language systematically to understand advantages and disadvantages in terms of continual training, instruction fine-tuning, task-specific fine-tuning, and vocabulary extension, (ii) investigate model size and multilingual models for adaptation, and (iii) publish all resources including source codes, datasets, and generative models reported in the experiments\footnote{https://github.com/metunlp/llamaturk}.

\section{Related Work}

Generative LLMs are either proprietary or open-source. Although proprietary LLMs have currently outstanding performance \cite{sun2024trustllm}, there are also efforts to develop competitive open-source models \cite{touvron2023llama, jiang2023mistral}.

The majority language of open-source generative LLMs is English. Their pretraining text corpus mostly includes text in the English language. For adapting LLMs pretrained with English data for low-resource languages, the following methods are examined. (i) The training phase is continued using non-English raw data to learn the language properties of the new language \cite{larcher2023cabrita,cui2024efficient,zhao2024llama,acikgoz2024bridging}. (ii) The knowledge of large language model is transferred by supervised fine-tuning on a non-English instruction or downstream-task dataset \cite{santilli2023camoscio,holmstrom-doostmohammadi-2023-making,kohli2023building,zhao2024llama,garcia2024introducing,kuulmets2024teaching}. (iii) The vocabulary of large language model is extended to include non-English tokens \cite{cui2023efficient,zhao2024llama}. 

These methods are employed in different studies and languages, resulting in a lack of understanding advantages and disadvantages of each in a controlled experimental framework.
Different from these studies, we provide a comprehensive experimental setup on the benefits of different methodologies for adapting generative LLMs for low-resource languages. Moreover, we focus on model size and multilingual models for adaptation.

\section{Adaptation Methods}
\label{sec:methods}
In this section, we explain the methods to adapt open-source generative LLMs for low-resource languages in detail. 

    \subsection{Continual Training}
    \label{sec:continual}
    Continual training is the process of extending the pretraining phase of LLMs by incorporating new data corpus \cite{gupta2023continual}. The main objective is to minimize the loss on this new data while having relatively lower loss scores on previous data since continual training is open to catastrophic forgetting \cite{french1999catastrophic,li2024examining}. Continual training can therefore capture implicit language structures and text semantics.
    
    Previous studies \cite{qin2022elle} show that continual training improves the performance of domain adaptation for BERT-like encoder-based LLMs \cite{devlin2019bert}. It is also used for adapting decoder-based generative LLMs to low-resource \cite{cui2023efficient, zhao2024llama}, code-mixed \cite{owen2024komodo}, non-Latin \cite{husain2024romansetu}, and multilingual \cite{lin2024mala} settings. 
    
    In this study, similar to previous studies, we employ Low-Rank Adaptation (LoRA) \cite{hu2021lora} for efficient training due to limited resources. We use a raw Wikipedia corpus\footnote{https://huggingface.co/datasets/wikipedia} from November 2023 with a size of 534,988 Turkish articles. 
    
    We set the input sequence length as 512 tokens and the batch size as 128 instances. We use 32 gradient accumulation steps and 100 linear warmup steps. We train with a learning rate of 3e-4 for a single epoch. LoRA's R is set to 8, alpha to 16, and dropout to 0.05. Since continual training is costly and the study has a limited budget, we employ continual training for only Llama-7b\footnote{https://huggingface.co/huggyllama/llama-7b} with 8-bit quantization. A single run of continual training takes approximately 206 hours with these settings using four NVIDIA RTX A4000s.
    
    \subsection{Instruction Fine-tuning}
    \label{sec:instruction}
    Instruction tuning is a supervised fine-tuning method that improves the ability of LLMs to follow instructions \cite{wei2021finetuned,ouyang2022training,zhang2024instruction}. During training, the model is presented with many pairs of instructions and corresponding responses. The main objective is to teach the model to generate accurate responses based on the given instructions, rather than continuing from the previous text.

    Different from previous instruction-tuning efforts, Stanford's Alpaca \cite{alpaca} is a leading model that shows major improvements by instruction fine-tuning an open-source generative LLM, namely  \cite{touvron2023llama}. While Alpaca and similar models such as Vicuna \cite{vicuna2023} have an instruction set constructed by prompting proprietary LLMs, other models such as Dolly \cite{DatabricksBlog2023DollyV2} employ human labor for constructing a more reliable instruction set. The majority of these efforts are for the English language, yet there are instruction-tuned models to adapt English-supported LLMs for low-resource settings \cite{cui2023efficient,zhao2024llama,azime2024enhancing}. 
    
    In this study, we construct an instruction set by translating Alpaca's 52k instructions from English to Turkish by using Google Translate\footnote{https://translate.google.com}. The quality of the translated set is inadequate for training since we observe many issues such as translation errors (e.g. missing letters and untranslated words), keyword translations (e.g. reserved keywords specific to programming languages should not be translated), and semantic mismatching (e.g. original instruction asks for a phrase with five words, but correct translation has less than five words). We therefore manually validate and correct the quality of the instruction set. We publish our instruction set\footnote{https://github.com/metunlp/llamaturk}. We also provide a prompting example for instruction fine-tuning in Appendix \ref{sec:appendix_prompt_instruction}.

    We employ instruction tuning for all LLMs examined in this study, namely Llama-7b\footnote{https://huggingface.co/huggyllama/llama-7b}, Llama-13b\footnote{https://huggingface.co/huggyllama/llama-13b}, and MaLA-10b\footnote{https://huggingface.co/MaLA-LM/mala-500-10b-v1}. We use 8-bit quantization with LoRA (resulting in training 12.4\% of LLM parameters) and the same hyperparameters as in continual training, except that we use a smaller input sequence length (256 tokens) and train for two epochs. A single run of instruction tuning takes approximately 17.5 hours for Llama-7b with these settings using two NVIDIA RTX 2080Tis.
    
    \subsection{Task-Specific Fine-tuning}
    \label{sec:task}
    Task-specific tuning is a type of instruction tuning, where a fine-tuning set involves task-related instructions and ground-truth answers \cite{budzianowski2019hello,wang2024far}, rather than adapting a general-purpose instruction set. Task-specific tuning of generative LLMs is proven to be successful in different domains including text editing \cite{raheja2023coedit}, sentiment analysis \cite{inserte2024large}, and machine translation \cite{zheng2024finetuning}. However, task-specific tuning have the potential of deteriorating the language capabilities of LLMs \cite{zhang2023balancing,zhao2023domain}. 
    
    We follow instruction fine-tuning with a task-specific dataset for the downstream task of sentiment analysis. We choose sentiment analysis since it is a widely applicable task that represents a fundamental natural language processing capability \cite{10.5555/3019323}. For this purpose, we create an instruction set for sentiment analysis. To create a balanced set, we downsample 2,500 instances for both negative and positive sentiment classes, a total of 5k instances from the TRSAv1 dataset \cite{aydougan2023trsav1}. We then use a prompt manually crafted for the task of sentiment analysis\footnote{We run prompts from existing resources \cite{bach2022promptsource} but decided to use a manually crafted one by observing better performance in preliminary experiments.}. We provide the prompt in Appendix \ref{sec:appendix_prompt_task}.

    We employ task-specific tuning for all LLMs examined in this study. We use all models in 8-bit quantization. We also use LoRA (resulting in training 12.4\% of LLM parameters) and the same hyperparameters as in instruction tuning. A single run of task-specific tuning takes approximately 2.5 hours for Llama-7b with these settings using two NVIDIA RTX 2080Tis.

\begin{table}[t]
\small
\begin{tabular}{llrr}
\hline
\textbf{} & \textbf{Data} & \textbf{Size} & \textbf{Tokens} \\
Continual training & Wiki & 535.0k & 273.9m \\
Instruction tuning & Alpaca & 52.0k & 13.3m \\
Task-specific tuning & Sentiment & 5.0k & 1.3m \\
Vocabulary extension & BPE & 28.6k & 28.6k \\
\hline
\end{tabular}
\caption{\textbf{Data statistics for adaptation methods}. The columns represent the type of data used (Data), the total number of instances (Size), and the total number of tokens (Tokens), respectively.}
\label{tab:datastats}
\end{table}

    \subsection{Vocabulary Extension}
    \label{sec:vocab}
    Vocabulary embeddings are a major component of how LLMs understand and process natural language text by capturing semantic meanings and relationships among subwords called tokens \cite{toraman2023impact}. Vocabulary tokens are determined by tokenization algorithms such as WordPiece \cite{6289079} and Bype Pair Encoding (BPE) \cite{sennrich-etal-2016-neural}.
    
    Llama has a vocabulary size of 32k tokens based on BPE tokenization \cite{touvron2023llama}. The majority of tokens in its vocabulary are English. The remaining small portion involves European languages with Latin and Cyrillic symbols. 

    In this study, we extend Llama's vocabulary by merging with low-resource language tokens. Specifically, we use the Turkish tokenizer with 28,600 tokens trained by BPE algorithm \cite{toraman2023impact} (We publish the tokenizer\footnotemark[6]). 

    Merging the original Llama tokenizer with low-resource vocabulary yields 59,773 tokens, meaning that 827 tokens are overlapping. This results in adding almost 228m new parameters to be trained into the model due to the extended vocabulary embeddings. We employ vocabulary extension with above-mentioned methods when Llama-7b is used with LoRA due to limited resources.

    \subsection{Combinations}
    A summary of data statistics used for the adaptation methods is given in Table \ref{tab:datastats}. In addition to a single examination of these methods, we also report the results of using them in combination to leverage better performance. We particularly employ the following combinations using Llama-7b with LoRA. Hyperparameters are set the same as explained in the previous subsections.

    \emph{Continual Training with Instruction Fine-tuning:} We first obtain a model by continual training using raw Wiki data as explained in Section \ref{sec:continual}. We then apply instruction fine-tuning as explained in Section \ref{sec:instruction}. The motivation is to boost the potential of instruction tuning when the backbone model is trained with low-resource raw text beforehand.
    
    \emph{Continual Training with Task-Specific Fine-tuning:} With a similar motivation to the previous approach, we first obtain a model by continual training using raw Wiki data. We then apply task-specific fine-tuning as explained in Section \ref{sec:task}.
    
    \emph{Continual Training with Instruction and Task-Specific Fine-tuning:} The motivation is to boost the performance of task-specific tuning when the model is trained by both raw text and instruction-set in low-resource language beforehand. We first obtain a model by continual training using raw Wiki data. We then apply instruction tuning and task-specific fine-tuning respectively. 
    
    \emph{Instruction and Task-Specific Fine-tuning:} This approach avoids continual training but examines using both instruction and then task-specific tuning respectively. The motivation is to boost the performance of task-specific tuning when the model is trained by only instruction-set in low-resource language beforehand.

    \emph{Vocabulary Extension with Instruction Fine-tuning:} We extend the vocabulary with low-resource language tokens as explained in Section \ref{sec:vocab}. We then apply instruction tuning to understand the impact of vocabulary extension on instruction tuning.

    \emph{Vocabulary Extension with Task-Specific Fine-tuning:} With a similar motivation to the previous approach, we extend the vocabulary with low-resource language tokens and then apply task-specific tuning to understand the impact of vocabulary extension on task-specific tuning.

    \emph{Vocabulary Extension with Continual Training:} We extend the vocabulary with low-resource language tokens and then apply continual training to understand its impact on continual training.

\section{Experiments}
In this section, we evaluate the performance of different methods to adapt generative large language models for low-resource language. We particularly conduct both intrinsic and extrinsic evaluations in order to understand the performance of both language comprehension and downstream tasks. We also run benchmark LLM evaluation by using appropriate datasets. This section further involves the results of using varying model sizes and applying multilingual models for the adaptation.

\subsection{Intrinsic Evaluation}
Intrinsic evaluation of generative LLMs involves a perplexity score that represents how well a language model can predict the next word in a sequence of text \cite{10.5555/1214993}:

\begin{equation}
\text{perplexity} = 2^{-\frac{1}{N}\sum_{i=1}^N \log_2 P(w_i|w_1, \dots, w_{i-1})}
\end{equation}

\noindent where $N$ is the total number of words and $P(w_i|w_1, w_2, \dots, w_{i-1})$ is the probability assigned by the model to the $i$-th word given the preceding text context. 

A lower perplexity score indicates that language model is better able to predict the next word, and thus has a better understanding of the language. 

\begin{table}[t]
\centering
\small
\begin{tabular}{lrrrr}
\hline
\textbf{} & \multicolumn{1}{c}{\textbf{xquad}} & \multicolumn{1}{c}{\textbf{xquad}} & \multicolumn{1}{c}{\textbf{dbricks}} & \multicolumn{1}{c}{\textbf{dbricks}} \\
\textbf{} & \textbf{question} & \textbf{context} & \textbf{instruction} & \textbf{response} \\
%Instances & 1,190 & 1,190 & 15,014 & 15,014 \\
Size & 1.2k & 1.2k & 15.0k & 15.0k \\
%Characters & 74,671 & 965,351 & 1,126,323 & 5,409,825 \\
Chars & 74.7k & 965.4k & 1.1m & 5.4m \\
%Tokens & 37,430 & 458,348 & 549,824 & 2,400,320 \\
Tokens & 37.4k & 458.3k & 549.8k & 2.4m \\
\hline
\end{tabular}
\caption{\textbf{Dataset statistics for perplexity}. The xquad dataset has question and context subsets. The databricks (dbricks) dataset has instruction and response subsets.}
\label{tab:perpstats}
\end{table}

\begin{table*}[t]
\centering
\small
\setlength\tabcolsep{0pt}
\begin{tabular*}{\linewidth}{@{\extracolsep{\fill}}lcccc|rrrr}
\hline
\multirow{3}{*}{\textbf{Model}}  & \multirow{2}{*}{\textbf{Continual}} & \multirow{2}{*}{\textbf{Instruction}} & \multirow{2}{*}{\textbf{Task}} & \multirow{2}{*}{\textbf{Vocabulary}} & \multicolumn{4}{c}{\textbf{Data}} \\

& \multirow{2}{*}{\textbf{Training}} & \multirow{2}{*}{\textbf{Tuning} } & \multirow{2}{*}{\textbf{Tuning}} & \multirow{2}{*}{\textbf{Extension}} & \multicolumn{1}{c}{\textbf{xquad}} & \multicolumn{1}{c}{\textbf{xquad}} & \multicolumn{1}{c}{\textbf{dbricks}} & \multicolumn{1}{c}{\textbf{dbricks}} \\
& &  &  &  & \textbf{question} & \textbf{context} & \textbf{instruction} & \textbf{response} \\
\texttt{Llama-7b} &  &  &  & & 6.6916 & 1.5487 & 9.5845 & 9.0259 \\
\texttt{LlamaTurk-7b-c} & \usym{1F5F8} &  & & & \textbf{5.5088} & \textbf{1.5064} & 8.4364 & \textbf{7.0924} \\
\texttt{LlamaTurk-7b-i} &  & \usym{1F5F8} & & & 6.3260 & 1.5674 & 8.3131 & 7.9351 \\
\texttt{LlamaTurk-7b-t} &  &  & \usym{1F5F8} & & 9.2267 & 1.7850 & 13.7173 & 13.2289 \\
\texttt{LlamaTurk-7b-c-i} & \usym{1F5F8} & \usym{1F5F8} & & & 7.0676 & 1.5978 & \textbf{8.2488} & 9.4570 \\
\texttt{LlamaTurk-7b-i-t} &  & \usym{1F5F8} & \usym{1F5F8} & & 9.0380 & 1.8194 & 13.0113 & 11.8501 \\
\texttt{LlamaTurk-7b-c-t} & \usym{1F5F8} &  & \usym{1F5F8} & & 7.7305 & 1.7181 & 12.5591 & 10.7188 \\
\texttt{LlamaTurk-7b-c-i-t} & \usym{1F5F8} & \usym{1F5F8} & \usym{1F5F8} & & 8.0855 & 1.6666 & 11.5441 & 10.6943 \\
\texttt{LlamaTurk-7b-v-i} & & \usym{1F5F8} & & \usym{1F5F8} & 18.6241 & 3.8897 & 22.1750 & 24.3312 \\
\texttt{LlamaTurk-7b-v-t} & & & \usym{1F5F8} & \usym{1F5F8} & 28.7707 & 5.8666 & 37.6394 & 43.7040 \\
\texttt{LlamaTurk-7b-v-c} & \usym{1F5F8} & & & \usym{1F5F8} & 17.3135 & 3.6807 & 23.9212 & 23.2612 \\
\hline
\end{tabular*}
\caption{\textbf{Perplexity scores}. The models have different adaptation methods: Continual Training (\texttt{c}), Instruction Tuning (\texttt{i}), Task-specific Tuning (\texttt{t}), and Vocabulary Extension (\texttt{v}). The xquad dataset has question and context subsets. The databricks (dbricks) dataset has instruction and response subsets. The best (lowest) perplexity scores for each dataset are given in bold.}
\label{tab:perplexity}
\end{table*}

We calculate the perplexity scores on different data collections than the ones used in Section \ref{sec:methods}. Specifically, we use the Turkish question and context subsets of \texttt{xquad} \cite{Artetxe:etal:2019}, and the instruction and response subsets of \texttt{databricks-dolly-15k} \cite{DatabricksBlog2023DollyV2} using a Turkish translated version\footnote{https://huggingface.co/datasets/atasoglu/databricks-dolly-15k-tr}. The detailed statistics of the data used for calculating perplexity scores are given in Table \ref{tab:perpstats}. The reason for reporting the perplexity scores for different subsets is that the characteristics of each subset can be helpful to understand the applied method's impact on the adaptation. For instance, \texttt{xquad-question} has instances of questions while \texttt{xquad-context} has longer paragraphs of task descriptions. Similarly, \texttt{databricks-instruction} has instruction-type questions, while \texttt{databricks-response} has answers or responses to those questions.

In Table \ref{tab:perplexity}, we provide the perplexity scores. The main observations can be summarized as follows.

\paragraph{Continual training reduces perplexity scores.} In all cases, perplexity scores are improved by continual training (\texttt{LlamaTurk-7b-c}). The lowest perplexity scores are also obtained by continual training in the majority of cases (three of four data collections). A possible reason is that the model could gradually accumulate language knowledge as it is exposed to more raw text. This incremental learning process can allow the model to become more robust and adaptable.

\paragraph{Instruction tuning improves perplexity but not task-specific tuning.} Perplexity scores are improved by instruction tuning (\texttt{LlamaTurk-7b-i}). The only exception is \texttt{xquad-context}, yer instruction tuning has still a very close perplexity score to the original \texttt{Llama-7b}. Our instruction-tuning set is based on Alpaca, which has general-purpose instructions and responses. On the other hand, task-specific tuning (\texttt{LlamaTurk-7b-t}) deteriorates perplexity scores in all cases. We argue that, by training on task-specific instructions, generative LLMs might become overly specialized and optimized for those specific instructions, rather than maintaining a more general understanding of language. 

\paragraph{Combinations fail in most cases but depends on data types.} The combinations that include task-specific tuning have poor perplexity scores. On the other hand, continual training and instruction tuning improve perplexity. We therefore expect to have a better performance by using them together (\texttt{LlamaTurk-7b-c-i}) but perplexity scores get worse than the case when they are applied alone. However, when perplexity is measured on an instruction set (\texttt{databricks-instruction}), continual training together with instruction tuning has the lowest perplexity score. This observation can support that generative LLMs adapt to different data types, and one should consider target data type before selecting adaptation method.

\paragraph{Vocabulary extension has poor perplexity.} In all models where vocabulary extension is applied (\texttt{Llama-7b-v}), perplexity scores get higher than the original (\texttt{Llama-7b}). We argue that without sufficient training data and fine-tuning, the model can struggle to effectively incorporate the new vocabulary into its internal representations and learning processes. Similarly, \citep{zhao2024llama} observes negative impact of vocabulary extension, and also suggests that vocabulary extension might not be a suitable choice for small-scale continual training such as in our continual training with 0.2 billion tokens of the training data. Another reason could be the number of additional tokens in vocabulary (28k tokens), merged with the original tokenizer (32k tokens). More experimentation is needed to understand if a different number of new tokens in vocabulary works better in adaptation. 

\subsection{Extrinsic Evaluation}
\label{sec:extrinsic}
Generative LLMs employ human evaluations as an evaluation method to align with human judgments \cite{ouyang2022training}. However, human-based evaluation is labor-intensive, making it costly and less feasible for low-resource languages. On the other hand, LLM evaluation benchmarks offer reliable evaluation for downstream NLP tasks such as GLUE \cite{wang-etal-2018-glue} and SuperGLUE \cite{10.5555/3454287.3454581}. Similarly, there are evaluation frameworks and tools such as LM Evaluation Harness \cite{eval-harness} and MLflow\footnote{https://github.com/mlflow/mlflow}. However, they mostly support English benchmark datasets. Although multilingual datasets are published by some benchmarks, either they do not include the language used in this study, or the data size is small for task-specific tuning. We therefore craft an evaluation on sentiment analysis in this subsection\footnote{We also provide a benchmark evaluation for available datasets from LLM benchmarks in Section \ref{sec:benchmark}.}.

\begin{table}[t]
\centering
\small
\setlength\tabcolsep{0pt}
\begin{tabular*}{\linewidth}{@{\extracolsep{\fill}}lcccc}
\hline
\textbf{Model} & \textbf{0-shot} & \textbf{1-shot} & \textbf{2-shot} & \textbf{3-shot}  \\
\texttt{Llama-7b} & \cellcolor[HTML]{e6fff5}0.00 & \cellcolor[HTML]{e6fff5}0.50 & \cellcolor[HTML]{66FFCC}0.53 & 0.50 \\
\texttt{LlamaTurk-7b-c} & \cellcolor[HTML]{e6fff5}0.00 & \cellcolor[HTML]{e6fff5}0.47 & \cellcolor[HTML]{66FFCC}0.54 & \cellcolor[HTML]{66FFCC}0.51 \\
\texttt{LlamaTurk-7b-i} & \cellcolor[HTML]{e6fff5}0.06 & \cellcolor[HTML]{e6fff5}0.48 & \cellcolor[HTML]{e6fff5}0.48 & \cellcolor[HTML]{66FFCC}0.56 \\
\texttt{LlamaTurk-7b-t} & \cellcolor[HTML]{398564}0.90 & \cellcolor[HTML]{419873}0.84 & \cellcolor[HTML]{52bf90}0.61 & \cellcolor[HTML]{49ab81}0.78 \\
\texttt{LlamaTurk-7b-c-i} & \cellcolor[HTML]{e6fff5}0.10 & \cellcolor[HTML]{66FFCC}0.52 & \cellcolor[HTML]{e6fff5}0.50 & \cellcolor[HTML]{66FFCC}0.54 \\
\texttt{LlamaTurk-7b-i-t} & \cellcolor[HTML]{419873}0.83 & \cellcolor[HTML]{398564}0.90 & \cellcolor[HTML]{317256}0.93 & \cellcolor[HTML]{398564}0.89 \\
\texttt{LlamaTurk-7b-c-t} & \cellcolor[HTML]{419873}0.82 & \cellcolor[HTML]{52bf90}0.60 & \cellcolor[HTML]{52bf90}0.62 & \cellcolor[HTML]{419873}0.86 \\
\texttt{LlamaTurk-7b-c-i-t} & \cellcolor[HTML]{52bf90}0.62 & \cellcolor[HTML]{66FFCC}0.52 & \cellcolor[HTML]{66FFCC}0.56 & \cellcolor[HTML]{66FFCC}0.51 \\
\texttt{LlamaTurk-7b-v-i} & \cellcolor[HTML]{e6fff5}0.35 & \cellcolor[HTML]{e6fff5}0.44 & \cellcolor[HTML]{e6fff5}0.49 & \cellcolor[HTML]{66FFCC}0.53  \\
\texttt{LlamaTurk-7b-v-t} & \cellcolor[HTML]{e6fff5}0.44 & \cellcolor[HTML]{e6fff5}0.50 & \cellcolor[HTML]{66FFCC}0.53 & \cellcolor[HTML]{66FFCC}0.53 \\
\hline
\end{tabular*}
\caption{\textbf{Accuracy scores on sentiment analysis}.
The darker cell color gets, the better task performance.}
\label{tab:sentiment}
\end{table}

\begin{table*}[t]
\centering
\small
\begin{tabular}{lcccc|cccc}
\hline
\multirow{2}{*}{\textbf{Model}} & \multicolumn{4}{c|}{\textbf{XCOPA}} & \multicolumn{4}{c}{\textbf{Belebele}}\\
 & \textbf{0-shot} & \textbf{1-shot} & \textbf{2-shot} & \textbf{3-shot} & \textbf{0-shot} & \textbf{1-shot} & \textbf{2-shot} & \textbf{3-shot}  \\
Llama-7b & 0.53 & 0.51 & 0.48 & 0.52 & 0.23 & 0.23 & 0.23 & 0.24 \\
LlamaTurk-7b-i & \textbf{0.58} & 0.51 & 0.50 & 0.55 & 0.24 & 0.27 & 0.25 & \textbf{0.28} \\
LlamaTurk-7b-c-i & 0.52 & 0.52 & 0.53 & 0.50 & 0.24 & 0.25 & 0.23 & 0.27 \\
LlamaTurk-7b-v-i & 0.55 & 0.53 & 0.54 & 0.54 & 0.24 & 0.27 & 0.23 & \textbf{0.28}  \\
\hline
\end{tabular}
\caption{\textbf{Accuracy scores on benchmark datasets}. The highest scores for each dataset are given in bold.}
\label{tab:benchmarks}
\end{table*}

For this purpose, we extract 100 instances (50 instances for both positive and negative classes) from the Turkish sentiment analysis dataset used in task-specific tuning \cite{aydougan2023trsav1}. We avoid selecting from 5k instances used in task-specific tuning explained in Section \ref{sec:task}. Since inference is time costly, we use a small subset of this dataset for the evaluation. We also craft inference prompts for different scenarios including zero-shot to few-shot prompts. We check the generated text if it equals to positive or negative, and calculate the accuracy score accordingly. We measure accuracy since the inference dataset is fully balanced. We provide the inference prompts in Appendix \ref{sec:appendix_prompt_task_inference}.

During inference, we load the models with 8-bit quantization due to limited hardware. Generation configuration involves the following hyperparameters. The temperature is set to 0.2. Beam search is applied with four beams, and top-p is set to 0.75. A single run of inference takes approximately from six hours (zero-shot) to eight hours (3-shot) for Llama-7b with these settings using two NVIDIA RTX 2080Tis.

In Table \ref{tab:sentiment}, we provide the perplexity scores for all methods. The main observations are as follows.

\paragraph{Task-specific tuning improves the performance of downstream task.} We find that task-specific tuning cannot help improve perplexity scores previously. However, our extrinsic evaluation shows that task-specific tuning improves the performance of sentiment analysis. Specifically, we observe that task-specific tuned model (\texttt{LllamaTurk-7b-t}) is good at zero-shot inference, suggesting that task-specific instructions provide sufficient knowledge for zero-shot evaluation. 

\paragraph{Instruction tuning boosts the performance of downstream task when used together with task-specific tuning.} When instruction tuning is employed alone, it has no significant impact on the performance of downstream task. However, we find that the highest accuracy score is obtained when instruction tuning and task-specific tuning are together employed (\texttt{LllamaTurk-7b-i-t}). Moreover, \texttt{LllamaTurk-7b-i-t} has a better few-shot performance compared to other methods including task-specific tuning. 

\paragraph{Continual training can help task-tuning.} When continual training is employed alone (\texttt{LllamaTurk-7b-c}), we observe no significant improvement in the performance of downstream task. However, the performance is promising when it is used together with task-specific tuning (\texttt{LllamaTurk-7b-c-t}). This suggests further examination of continual training with task-specific tuning in different downstream tasks and datasets.

\paragraph{Vocabulary extension has poor downstream performance.} Similar to the perplexity experiments, we observe that vocabulary extension has no improvement on the performance of downstream task.

\subsection{Benchmark Evaluation}
\label{sec:benchmark}
In this subsection, we report the performance results on benchmark datasets. Since LLM evaluation benchmarks mostly include English datasets, we examine multilingual datasets in available LLM benchmarks. For this purpose, we use the Turkish subsets of XCOPA \cite{ponti2020xcopa} and Belebele \cite{bandarkar2023belebele} datasets provided by LM Evaluation Harness \cite{eval-harness}. XCOPA is a benchmark to evaluate the ability of machine learning models to transfer commonsense reasoning. Belebele is a multiple-choice machine reading comprehension dataset, and each question has four multiple-choice. We modify the default prompts given in LM Evaluation Harness to align with our instruction prompting. We provide the inference prompts in Appendix \ref{sec:appendix_prompt_xcopa_inference} and \ref{sec:appendix_prompt_belebele_inference}.

Since the dataset sizes are small, we are not able to apply task-specific tuning in these benchmark datasets. Specifically, we observe almost no change in performance scores when XCOPA's 600 and Belebele's 900 instances are fine-tuned for the Turkish language, while the performance is improved in Section \ref{sec:extrinsic} with 5k instances. We thereby report the results for instruction tuning and related methods. Table \ref{tab:benchmarks} reports the accuracy scores on the XCOPA and Belebele datasets. 

The results show that instruction tuning (\texttt{LlamaTurk-7b-i}) improves the performance of downstream task in both datasets. However, continual training and vocabulary extension have no significant benefits on the results. The results thereby align with the results of sentiment analysis reported in Section \ref{sec:extrinsic}.

\subsection{Model Size}
\begin{figure}[t]
 \begin{subfigure}{0.23\textwidth}
     \includegraphics[width=\textwidth]{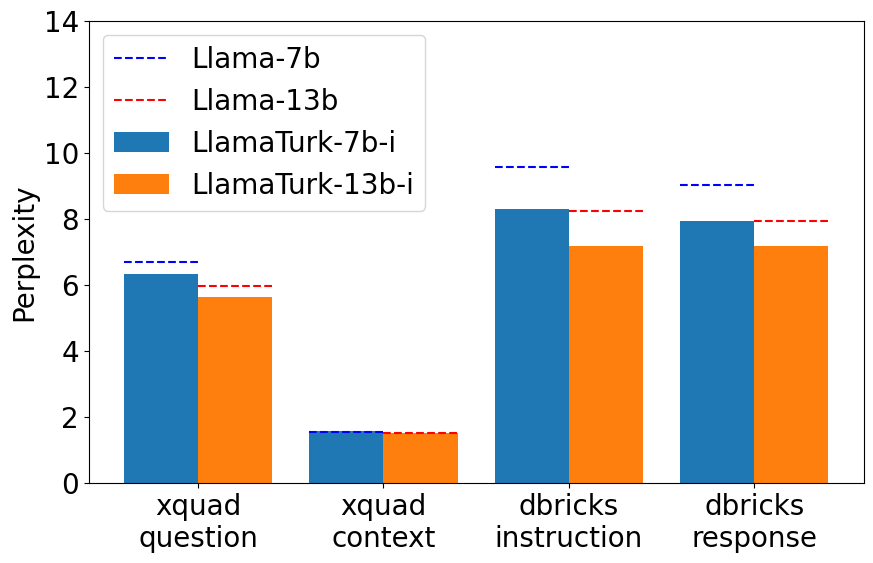}
     \caption{Perplexity of instruction tuning (lower is better)}
     \label{fig:a}
 \end{subfigure}
 %\hfill
 \begin{subfigure}{0.23\textwidth}
     \includegraphics[width=\textwidth]{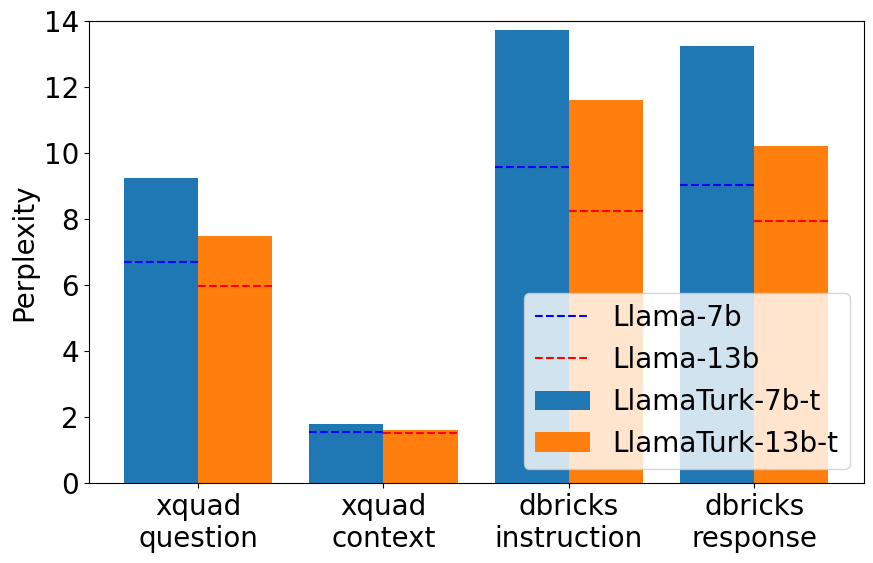}
     \caption{Perplexity of task-specific tuning (lower is better)}
     \label{fig:b}
 \end{subfigure}
 
 %\medskip
 \begin{subfigure}{0.23\textwidth}
     \includegraphics[width=\textwidth]{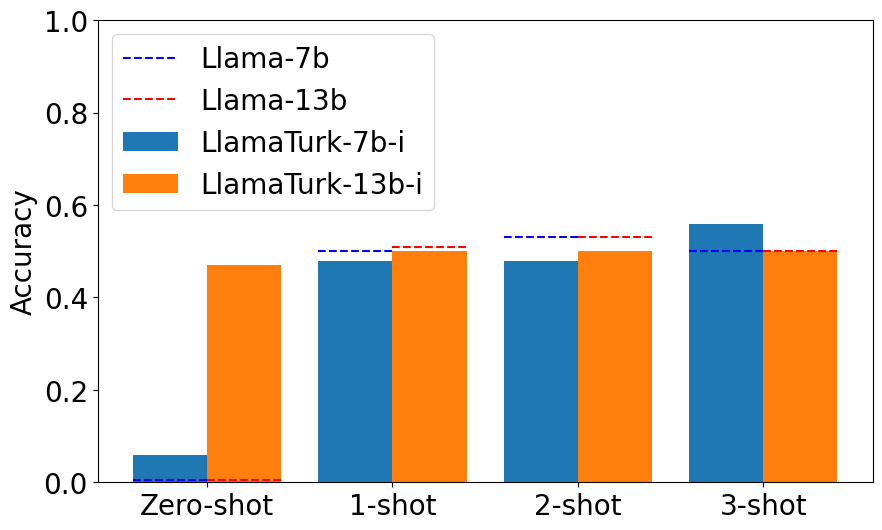}
     \caption{Accuracy of instruction tuning (higher is better)}
     \label{fig:c}
 \end{subfigure}
 %\hfill
 \begin{subfigure}{0.23\textwidth}
     \includegraphics[width=\textwidth]{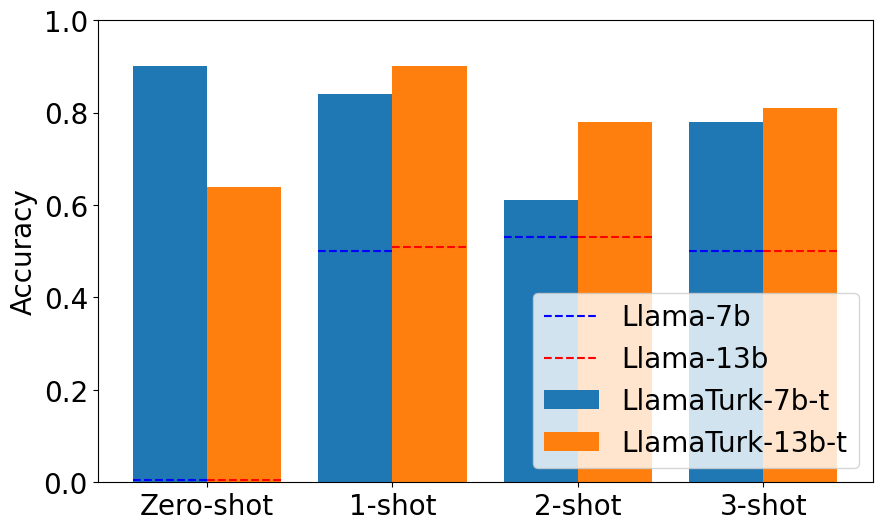}
     \caption{Accuracy of task-specific tuning (higher is better)}
     \label{fig:d}
 \end{subfigure}

 \caption{\textbf{Model size comparison for adaptation.}}
 \label{fig:model_size}
\end{figure}

We provide an analysis of the impact of model size on adapting generative LLMs. For this purpose, we employ Llama models with 7b and 13b parameters. Figure \ref{fig:model_size} shows a histogram depicting the comparison between the fine-tuned models for instruction tuning (\texttt{LlamaTurk-7b-i} and \texttt{LlamaTurk-13b-i}) and task-specific tuning (\texttt{LlamaTurk-7b-t} and \texttt{LlamaTurk-13b-t}).

\paragraph{Perplexity is improved by adapting a larger model.} In both cases of applying instruction or task-specific tuning, we find that \texttt{LlamaTurk-13b} improves perplexity scores in all cases. However, task-specific tuning (\texttt{LlamaTurk-13b-t}) is still outperformed by the original Llama model \texttt{Llama-13b} in most cases.

\paragraph{Task performance is improved by adapting a larger model when few-shot tuning is applied.} We find that \texttt{LlamaTurk-13b} improves the performance of downstream task when it is applied with task-specific tuning and few-shot evaluation. On the other hand, the adaptation of a larger model with instruction tuning has no significant impact on the performance of downstream task.

\subsection{Multilingual Models}
We also provide an analysis for the impact of multilingual generative LLMs on adapting generative LLMs. For this purpose, we fine-tune a multilingual model, MaLA-500 \cite{lin2024mala500}. MaLA is developed to cover 534 languages by using vocabulary extension and continual training on Llama2 \cite{touvron2023llama2}. Analyzing a multilingual LLM with an enriched vocabulary can provide more insights into LLM adaptation for low-resource languages.

\begin{figure}
 \begin{subfigure}{0.23\textwidth}
     \includegraphics[trim={0 17 0 0},width=\textwidth]{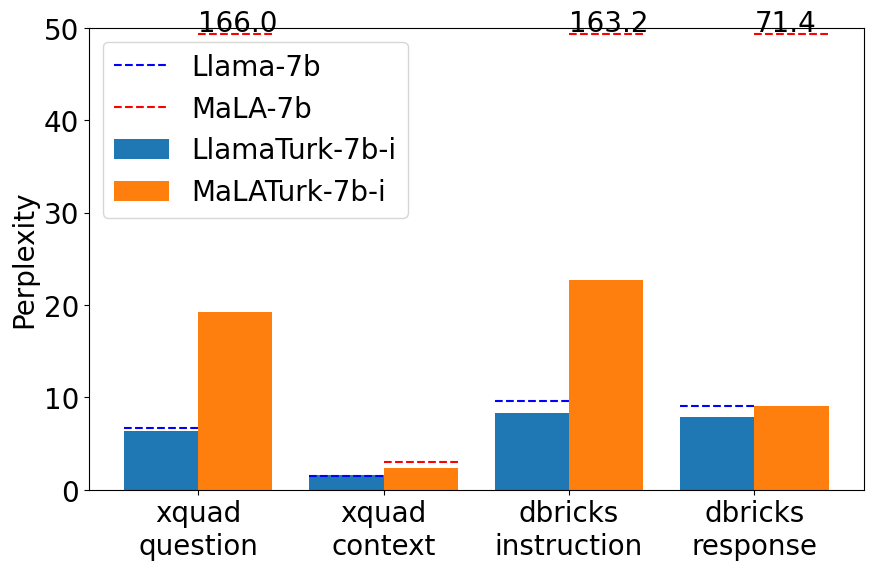}
     \caption{Perplexity of instruction tuning (lower is better).}
     \label{fig:a}
 \end{subfigure}
 \hfill
 \begin{subfigure}{0.23\textwidth}
     \includegraphics[trim={0 17 0 0},width=\textwidth]{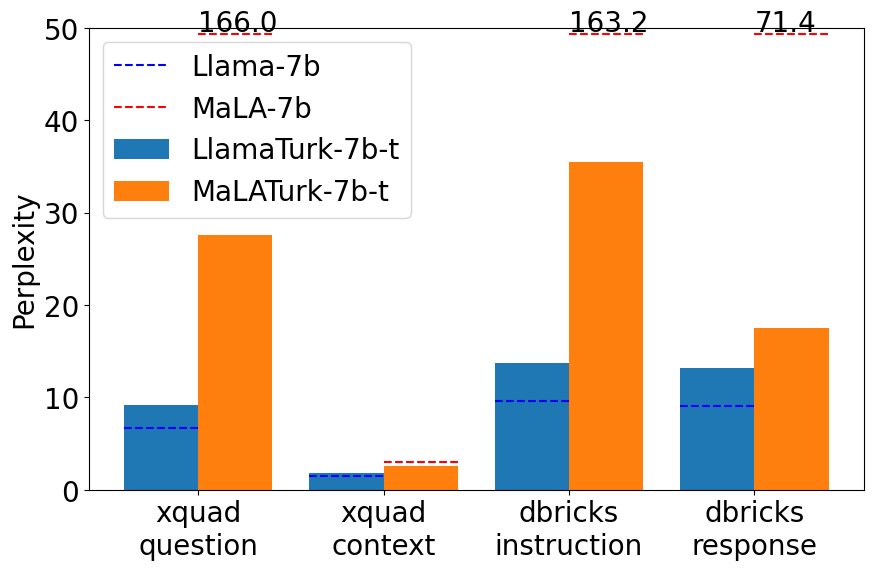}
     \caption{Perplexity of task-specific tuning (lower is better).}
     \label{fig:b}
 \end{subfigure}
 
 \medskip
 \begin{subfigure}{0.23\textwidth}
     \includegraphics[trim={0 10 0 12},width=\textwidth]{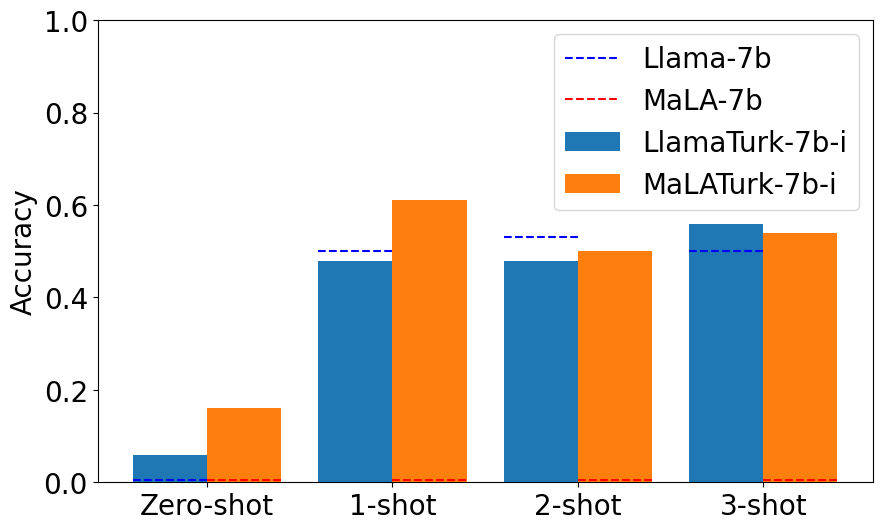}
     \caption{Accuracy of instruction tuning (higher is better).}
     \label{fig:c}
 \end{subfigure}
 \hfill
 \begin{subfigure}{0.23\textwidth}
     \includegraphics[trim={0 10 0 12},width=\textwidth]{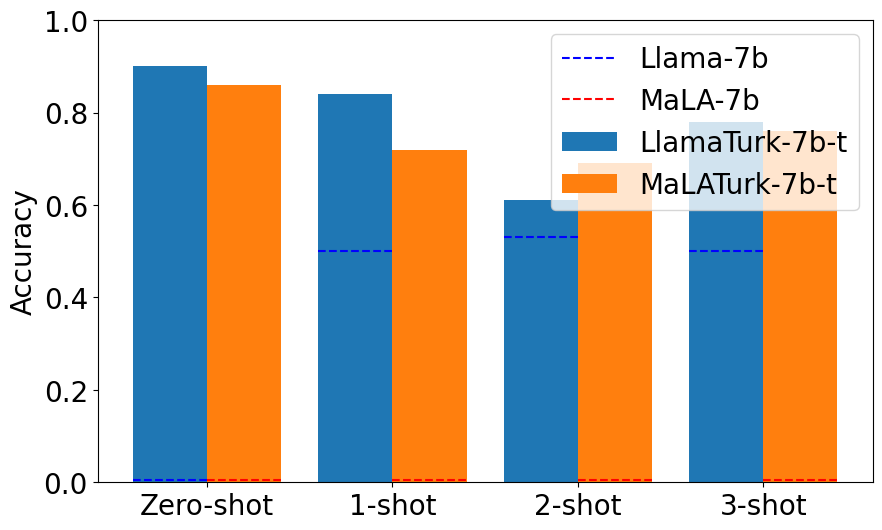}
     \caption{Accuracy of task-specific tuning (higher is better).}
     \label{fig:d}
 \end{subfigure}

 \caption{\textbf{Multilingual comparison for adaptation.}}
 \label{fig:multilingual}
\end{figure}

Figure \ref{fig:multilingual} shows a histogram depicting the comparison between the fine-tuned models for instruction tuning (\texttt{LlamaTurk-7b-i} and \texttt{MaLATurk-7b-i}) and task-specific tuning (\texttt{LlamaTurk-7b-t} and \texttt{MaLATurk-7b-t}).

\paragraph{Adapting multilingual LLM has no significant improvements.} Perplexity and accuracy scores of the original \texttt{MaLA-7b} model are improved by adapting \texttt{MaLATurk-7b} in both instruction and task-specific tuning. However, the perplexity of adapting a monolingual model \texttt{LlamaTurk-7b} is still better than adapting a multilingual model in all cases. Similarly, monolingual adaptation has better accuracy scores of task-specific tuning in most cases. The only benefit of adapting multilingual LLM is observed when instruction tuning is applied.

\section{Conclusion}
This study examines different methods for adapting English-dominant generative large language models to low-resource languages. 

The results show that continual training with raw text can improve perplexity, while vocabulary extension has no significant impact on adaptation performance. We also find that the adaptation with general-purpose instruction tuning has promising results in both perplexity and accuracy scores, while downstream task performance can be boosted by task-specific tuning. Furthermore, adapting a larger model with 13b parameters improves task performance with few-shot tuning. However, we observe no significant improvements by adapting a multilingual model. 

In future work, we plan to adapt other open-source language models such as Llama2 \cite{touvron2023llama2} and Gemini \cite{geminiteam2024gemini} to generalize our results to different models. Other adaptation methods can also be studied such as modification of model architecture since different model layers and tokenization algorithms might change the outcomes. 

\section{Limitations}
This study employs a particular family of generative large language models (Llama and MaLA) for adapting open-source generative monolingual and multilingual LLMs to a low-resource language. Using other generative models might have different results in the experiments. Similarly, we use the Turkish language for the target of adaptation. Other languages might have different experimental results depending on the tuning and inference datasets with prompt examples. We therefore acknowledge the effect of the instruction set and prompting templates in the results.

Moreover, benchmark evaluation is limited to multilingual datasets in this study due to the availability of benchmark datasets for the target language. Lastly, we would like to emphasize the limited hardware resources the experiments were conducted, which restricts using a variety of models including larger sizes (higher than 13b) and different model types (rather than Llama).

\section{Ethical Concerns}
This study employs a low-resource language, Turkish, and our findings can guide to other researchers studying low-resource languages. We also provide both intrinsic and extrinsic performance evaluations that can be considered for deploying generative LLMs in similar tasks. 

To provide transparency, we explain all details regarding text collections used in pretraining and fine-tuning our generative language models. Moreover, we report the details of the models and configurations with hyperparameters.

Since the training corpus of generative LLMs involves a huge amount of raw text from different resources including the world wide web, it is inevitable to observe a risk of cultural and ethical bias towards different individuals and communities in the generated text of the published models in this study \cite{kasneci2023chatgpt,cetinkaya}. Moreover, training texts are contaminated with more problematic biases and polluted with a large amount of synthetic text generated by LLMs \cite{denning2024can}. Possible bias can be removed by filtering the corpus, however, we leave the study of such filtering to future work since it would require a dedicated effort but the scope of this study is to compare the adaptation methods of generative LLMs for low-resource languages.

Lastly, we estimate the carbon footprint of our study based on the energy usage of GPUs. We consider execution time in hours and electrical energy consumption in kWh, and assume that power consumption during training is equal to the maximum power drain of GPUs by operating at maximum power utilization (0.25 MW for 2080Ti, and 0.14 MW for A4000). We assume that 1 MWh is equivalent to 0.439 ton CO2eq\footnote{https://enerji.gov.tr/evced-cevre-ve-iklim-elektrik-uretim-tuketim-emisyon-faktorleri}. Our estimation ignores the carbon footprint of CPU utilization and the manufacturing costs of the hardware. 

Social carbon cost is approximately 50.64, 3.84, and 0.55 kg CO2eq for a single run of continual training, instruction tuning, and task-specific tuning, respectively.

%For a single run of continual training by using four A4000s, social carbon cost is approximately 50.64 kg CO2eq (0.14 kW * 206 hours of training * 4 GPUs = 115.36 kWh). 

%For a single run of instruction tuning by using two 2080Tis, social carbon cost is approximately 3.84 kg CO2eq (0.25 kW * 17.5 hours of training * 2 GPUs = 8.75 kWh). 

%For a single run of task-specific tuning by using two 2080Tis, social carbon cost is approximately 0.55 kg CO2eq (0.25 kW * 2.5 hours of training * 2 GPUs = 1.25 kWh). 

\bibliography{custom}

\appendix

\section{Appendix}
\label{sec:appendix}

\subsection{Instruction Fine-tuning Prompt}
\label{sec:appendix_prompt_instruction}
The prompt used in instruction tuning is given as follows (translated prompt is given in parenthesis).

\begin{verbatim}
Aşağıda, daha geniş bir bağlam sağlayan 
girdiyle birlikte bir görevi açıklayan
talimat bulunmaktadır. Talimatı yeterince
sağlayan bir çıktı yaz. 
(Below is an instruction explaining a task 
with  input that provides more context. 
Write an output satisfying the instruction)

### Talimat (Instruction):
[INSTRUCTION]

### Girdi (Input):
[INPUT]

### Çıktı (Output):
[OUTPUT]
\end{verbatim}

\subsection{Task-Specific Fine-tuning Prompt}
\label{sec:appendix_prompt_task}
\noindent The prompt used in task-specific (sentiment analysis) fine-tuning is given as follows (translated prompt is given in parenthesis).

\begin{verbatim}
Aşağıda bir görevi açıklayan talimat 
bulunmaktadır. Talimatı yeterince 
sağlayan bir çıktı yaz.
(Below are instructions describing a task. 
Write an output that satisfying
the instruction)

### Talimat:
Lütfen verilen yorumun olumlu ya da 
olumsuz olduğunu çıktı olarak belirtin.
(Please indicate whether the given comment 
is positive or negative.)

### Yorum (Comment): 
[INPUT]

### Çıktı (Output):
[OUTPUT]
\end{verbatim}

\subsection{Task-Specific Inference Prompt}
\label{sec:appendix_prompt_task_inference}
For sentiment analysis, the prompt used in zero-shot inference is the same as the prompt used for task-specific fine-tuning given in \ref{sec:appendix_prompt_task}. Few-shot prompting (one-shot for example) is given as follows (translated prompt is given in parenthesis). 

\begin{verbatim}
Aşağıda bir görevi açıklayan talimat 
bulunmaktadır. Talimatı yeterince sağlayan 
bir çıktı yaz.
(Below are instructions describing a task. 
Write an output satisfying the instruction)
### Talimat (Instruction):
Lütfen verilen yorumun olumlu ya da 
olumsuz olduğunu çıktı olarak belirtin.
(Please indicate whether the given comment 
is positive or negative.)

### Yorum (Comment): 
çok güzel, sağlıklı, temiz, ferah 
(very beautiful, healthy, clean, spacious)

### Çıktı (Output):
olumlu
(positive)

### Talimat (Instruction):
Lütfen verilen yorumun olumlu ya da 
olumsuz olduğunu çıktı olarak belirtin.
(Please indicate whether the given comment 
is positive or negative.)

### Yorum (Comment): 
[INPUT]

### Çıktı (Output):
[OUTPUT]
\end{verbatim}

\subsection{XCOPA Inference Prompt}
\label{sec:appendix_prompt_xcopa_inference}
Few-shot prompting (one-shot for example) is given as follows (translated prompt is given in parenthesis). 

\begin{verbatim}
Aşağıda bir görevi açıklayan talimat 
bulunmaktadır. Talimatı yeterince 
sağlayan bir çıktı yaz.
(Below are instructions describing a task. 
Write an output satisfying the instruction)

### Talimat (Instruction): 
Verilen cümlenin sebebi nedir?
(What is the reason for the given sentence?)
Kadın kötü bir ruh halindeydi bu yüzden 
(The woman was in a bad mood so)

### Girdi (Input):
arkadaşıyla biraz konuştu. 
(she talked to her friend for a while.)
arkadaşına onu yalnız bırakmasını söyledi.
(she told her friend to leave her alone.)

### Çıktı (Output):
Kadın kötü bir ruh halindeydi bu yüzden 
arkadaşına onu yalnız bırakmasını söyledi.
(The woman was in a bad mood so she told
her friend to leave her alone.)

Aşağıda bir görevi açıklayan talimat 
bulunmaktadır. Talimatı yeterince sağlayan
bir çıktı yaz.
(Below are instructions describing a task. 
Write an output satisfying the instruction)

### Talimat (Instruction):
Verilen cümlenin sebebi nedir? 
(What is the reason for the given sentence?)
[INPUT] 

### Girdi (Input): 
[OPTION1]
[OPTION2]

### Çıktı (Output):
Ürün balonlu naylonla paketlenmişti 
bu yüzden [OUTPUT]
(The product was packaged with 
bubble wrap so [OUTPUT])
\end{verbatim}

\subsection{Belebele Inference Prompt}
\label{sec:appendix_prompt_belebele_inference}
Few-shot prompting (one-shot for example) is given as follows (translated prompt is given in parenthesis). 

\begin{verbatim}
Aşağıda bir görevi açıklayan talimat 
bulunmaktadır. Talimatı yeterince 
sağlayan bir çıktı yaz.
(Below are instructions describing a task. 
Write an output satisfying the instruction)

### Talimat (Instruction): 
Tüm notalara doğru şekilde basmaya devam 
ederken elinizin mümkün olduğu kadar 
rahat olduğundan emin olun - aynı zamanda
parmaklarınızla fazladan hareketler 
yapmamaya çalışın. Bu şekilde kendinizi
olabildiğince az yormuş olacaksınız. 
Unutmayın ki piyanoda olduğu gibi daha
fazla ses için tuşlara çok güçlü 
vurmanıza gerek yoktur. Akordeon 
üzerinde, ekstra hacim elde etmek için
körüğü daha fazla basınç veya hızda
kullanırsınız. Akordeonu çalarken 
aşağıdakilerden hangisi sesin 
yükselmesini sağlar?
(Make sure your hand is as relaxed as
possible while still hitting all the
notes correctly - at the same time,
try not to make extra movements with 
your fingers. This way, you will tire
yourself as little as possible. 
Remember that you don't need to hit 
the keys too hard to get more sound,
like on the piano. On the accordion,
you use the bellows with more pressure 
or speed to get extra volume.
Which of the following makes the sound
rise when playing the accordion?)

### Girdi (Input): 
A: Daha fazla hız (more speed) 
B: Daha fazla güç (more power)
C: Daha az basınç (less pressure)
D: Daha az parmak hareketi 
(less finger movement)

### Çıktı (Output):  
A

Aşağıda bir görevi açıklayan talimat 
bulunmaktadır. Talimatı yeterince 
sağlayan bir çıktı yaz.
(Below are instructions describing a task. 
Write an output satisfying the instruction)

### Talimat (Instruction): 
Tüm notalara doğru şekilde basmaya devam 
ederken elinizin mümkün olduğu kadar 
rahat olduğundan emin olun - aynı zamanda
parmaklarınızla fazladan hareketler 
yapmamaya çalışın. ... Akordeonu çalarken 
aşağıdakilerden hangisi sesin 
yükselmesini sağlar?
(Make sure your hand is as relaxed as
possible while still hitting all the
notes correctly - at the same time,
try not to make extra movements with 
your fingers. ... Which of the 
following makes the sound rise 
when playing the accordion?)

### Girdi (Input): 
[OPTION1]
[OPTION2]
[OPTION3]
[OPTION4]

### Çıktı (Output):
[OUTPUT]
\end{verbatim}

\end{document}